\documentclass[fleqn,10pt]{wlscirep}
\usepackage[utf8]{inputenc}
\usepackage[T1]{fontenc}

\usepackage{forest}
\usepackage{amssymb}
\usepackage{calrsfs}
\usepackage{graphicx}
\usepackage{colortbl}

\title{Uncertainty-aware Coarse-to-Fine Main Pancreatic Duct Segmentation for Width Measurement in 3D MRCP}

\author[1, *]{Zipei Zhang}
\author[1]{Jaakko Sahlsten}
\author[1]{Mikko Saukkoriipi}
\author[2,3]{Johanna Ronkainen}
\author[2,3]{Mona Lehto}
\author[2,3]{Akseli Aittoniemi}
\author[2,3]{Irina Rinta-Kiikka}
\author[1]{Kimmo Kaski}
\affil[1]{Department of Computer Science, Aalto University School of Science, Finland}
\affil[2]{Department of Radiology, Tampere University Hospital, Wellbeing Services County of Pirkanmaa, Finland}
\affil[3] {Faculty of Medicine and Health Technology, Tampere University, Finland}

\affil[*]{Corresponding author: zipei.zhang@aalto.fi}


\begin{abstract}
Assessment of the main pancreatic duct (MPD) from volumetric magnetic resonance cholangiopancreatography (MRCP) scans is an essential task for ductal dilatation diagnosis and pancreatic cancer analysis. However, manual assessment is affected by reader variability in which automated analysis may assist. Nevertheless, fully automated and uncertainty-aware MPD analysis from volumetric MRCP remains underexplored area of research. In this study, we propose an automatic and uncertainty-aware deep learning-based pipeline for MPD segmentation and maximum-width analysis from MRCP scans. A coarse localisation model first identifies the ductal region-of-interest, followed by a fine model that segments the MPD within the region. The dataset consists of 256 MPD-segmented MRCP scans from 69 patients. A randomly selected holdout test subset of 14 patients with 55 MRCP volumes was used for evaluation. We employed a five-model cross-validation ensemble for improved performance and uncertainty estimation. The best MPD segmentation performance was 0.634 DSC, 0.744 precision, and 0.593 recall. The mean absolute error for maximum MPD width prediction was 0.355~mm and dilated MPD ($\geq 5$~mm) was detected with 75.0\% sensitivity and 74.4\% specificity. These novel results suggest that automated MPD segmentation can be used to aid MRCP-based duct measurements.
\end{abstract}

\begin{document}
\flushbottom
\maketitle


\section{Introduction}
Pancreatic cancer, most commonly pancreatic ductal adenocarcinoma (PDAC), is one of the most aggressive cancers worldwide, ranking seventh among cancer-related deaths, with incidence still increasing~\cite{sung2021global}. One reason for the poor prognosis is that more than 80 \% of patients with PDAC are diagnosed in late stages which is why early 
detection is critical for improving prognosis, as tumors often become inoperable after infiltrating adjacent structures. 
Imaging plays a central role in identification of 
early disease, including small solid tumors and secondary morphological changes in the pancreatic parenchyma and main pancreatic duct (MPD). Computed tomography (CT) is the main imaging modality used in the diagnosis of 
PDAC, while 
magnetic resonance imaging (MRI) is used as a problem-solving imaging modality for cases that are not conclusive in CT. 

The main reason for this is that MRI provides a higher sensitivity for detecting small ($<2$ cm) isodense lesions compared to CT~\cite{choi2016comparison}. Another main advantage of MRI is the 
ability to obtain images of the ducts by magnetic resonance cholangiopancreatography (MRCP). MRCP imaging is based on heavily T2-weighted pulse sequences and thus highlights the natural differences in T2-weighted contrast between fluid-filled structures such as
ducts and surrounding soft tissues~\cite{griffin2012magnetic}. The MRCP is one of the 
imaging sequences taken routinely during abdominal MRI~\cite{huang2022standardization}.
In addition to an expansive tumour finding, imaging findings may be secondary~\cite{cannella2025conditions}. The most common of these is dilatation of the main pancreatic duct~\cite{kang2021factors,toshima2021ct}which 
can be associated with an obstructing tumour that causes
focal stricture with dilatation upstream, and the stricture is usually abrupt in morphology and over
6~mm long in malignant cases~\cite{kim2017isolated}. Changes in MPD can be detected in MRCP images~\cite{griffin2012magnetic}. 

In their study Toshima et al.~\cite{toshima2021ct} concluded 
that most patients with pancreatic ductal adenocarcinoma showed focal pancreatic abnormalities on CT performed at least 1 year before diagnosis. The focal change in MPD diameter exhibited the shortest duration from its development to subsequent diagnosis~\cite{toshima2021ct}. Hence, it is important to develop tools to maximise the acquisition of information about the morphology of the main duct.
Moreover, the volume and diameter of MPD as features have been shown to improve the 
classification performance of pancreatic ductal adenocarcinoma compared to using only 
radiomics features~\cite{wang2025predicting}.

Deep learning has been extensively applied in the field of medical image analysis, including a variety of tasks such as semantic segmentation, image to image enhancement, and classification. In medical image segmentation, U-net~\cite{ronneberger2015u}, i.e., the encoder-decoder architecture with skip connections, has become the standard model architecture. 
For volumetric segmentation, 3D variants such as 3D U-Net has been proposed~\cite{cciccek20163d}. More recently, complex variants of U-net have been proposed, such as nnU-Net~\cite{isensee2021nnu}, which include fully automated preprocessing pipelines and automated architectural choice, 
such as number of layers and features, and Swin-UNETR~\cite{hatamizadeh2021swin} that are still based on the encoder-decoder approach but instead of convolutional layers, it utilises shifted window approach and transformer blocks.  

The segmentation of the main
pancreatic duct 
using deep learning approaches has previously been studied using CT-based imaging. In the pancreatic domain, so far most automatic analysis of the main pancreatic duct (MPD) has been developed on CT rather than MRCP. Abi Nader et al.~\cite{abi2023automatic} proposed a deep learning pipeline on portal venous CT for pancreatic lesion detection and MPD dilatation assessment. Their method first segmented the pancreas, lesions, and visible MPD using nnU-Net, and then derived imaging features including MPD diameters in the pancreatic 
head, body, and tail. In a large multi-institutional cohort, they reported a mean MPD Dice score of 0.58 \(\pm\) 0.37, showing both the feasibility and difficulty of automatic duct segmentation. 
More recently, Tan et al.~\cite{tan2025pancreatic} studied MPD segmentation on photon-counting CT and reported an average Dice score of 0.621 using nnU-Net. In addition, Ambo et al.~\cite{ambo2025automatic} proposed an automated CT-based pipeline for MPD dilatation detection that combined pancreas segmentation, MPD segmentation, centreline estimation, and cross-sectional shape analysis. Together, these studies show that automatic duct analysis is possible, but the literature remains centred on CT.
MPD segmentation has been studied with high resolution, high contrast, and large field-of-view Photon-Counting CT scans 
using nnU-Net, which was trained using 53 scans and evaluated on 13 scans, resulting in average DSC of 0.621~\cite{tan2025pancreatic}.

In contrast to studies with CT, there are few studies on MPD analysis based on MRCP volumes, largely due to limited availability of 3D MRCP datasets with expert annotation. However, MRCP is clinically important for MPD evaluation due to its high contrast for fluid-filled structures without ionizing radiation. Existing MRCP-based studies in pancreatic imaging are more commonly focused on detection or characterisation of pancreatic cystic lesions, or they rely on manual or semi-automatic duct segmentation, rather than delivering a fully automated 3D MPD segmentation tool~\cite{shi2025deep}. Beyer et al.~\cite{beyer2023definition} established age-dependent reference values for the diameter of the pancreatic duct 
in MRCP in a large population dataset, showing that the diameter of the duct 
increases with age and that conventional upper limits may overestimate abnormality in healthy subjects. Ikezawa et al.~\cite{ikezawa2024correlation} studied the correlation between MPD diameter measurements on special pancreatic ultrasonography and MRCP, again treating MRCP as a measurement reference rather than as an input to automated image analysis. Similarly, Zhang et al.~\cite{zhang2023threshold} analysed MPD diameter on MRI to identify 
malignant intraductal papillary mucinous neoplasm, which reinforces 
the clinical importance of MPD width as a biomarker. Still there is no research work on the fully automated MPD segmentation-width analysis pipeline, the lack of which 
motivates the need for data-driven MRCP analysis approaches that explicitly address automated MPD width categorized classification.



\section{Methods}

In this study, we evaluated whether automatic segmentation can support quantitative maximum-width analysis of the main pancreatic duct from 3D MRCP volumes. We adopted established 3D encoder-decoder segmentation models in a task-specific coarse-to-fine pipeline designed for a thin tubular structure. Moreover, uncertainty of the model is quantified, analysed, and compared against its performance to estimate the utility of model uncertainty. 
The overall workflow consisted of three steps: (1) full-volume coarse localisation of the MPD region, (2) region-of-interest (ROI) extraction and detailed segmentation using a fine model, and (3) uncertainty estimation using five cross-validation models.

\subsection{Dataset, annotation and preprocessing}
\label{methods/dataset}

In this 
retrospective study, we used 256 MRCP volumes of 69 patients from Tampere University Hospital pancreatic cyst registry. The study was conducted in accordance with the Declaration of Helsinki, it was approved and the need for patients’ informed consent was waived by the Institutional Review Board of the Wellbeing Services County of Pirkanmaa, Finland (reference approval number: R24295). The patient records were retrospectively collected and pseudonymised by the data-providing authority prior to transfer to the research environment. 
All analyses were performed within a secure and access-controlled Aalto University's SECDATA computing environment compliant with national data protection policies and the General Data Protection Regulation (GDPR)~\cite{eu2016gdpr}.

\paragraph{Annotation procedure} 
Expert voxel-wise MPD segmentations were created in 3D Slicer (version 5.6.2)~\cite{fedorov20123d} primarily by two clinical researchers and after that, two experienced abdominal radiologists with over 20 years of MRI experience reviewed all the segmentation annotations. The annotation protocol focused on the visible main pancreatic duct in the source MRCP volume. Cases with uncertain visibility were reviewed jointly by the two radiologists to improve annotation consistency.

\paragraph{Image preprocessing} 
All MRCP volumes were resampled to a voxel spacing of \(0.9 \times 0.625 \times 0.625\,\mathrm{mm}^3\) and intensity-normalised on a per-volume basis before training. 

\paragraph{Dataset partitioning}
The dataset was divided 
at the patient level 
in the development and test sets with a 80:20 ratio, resulting in 201 and 55 scans, respectively. The characteristics of the dataset 
are shown in Table~\ref{tab:characteristics}. In model training, the development set was further divided 
into 5-folds, where four folds were used for the training subset and one fold as the validation data for model selection. The test dataset consisted of 55 scans from 14 patients having 1 to 7 scans each, and was used to evaluate the developed methods. 
In particular, for 
patients with multiple MRCP scans the test dataset, the mean interval between the earliest and latest scan was 2.89 years, with a standard deviation of 1.47 years. In the patients with multiple scans, each of the scans was taken 
at different time points. 


\begin{table}[h]
\centering
\caption{Characteristics of Development and Test Datasets. In our dataset, main pancreatic duct (MPD) dilation is caused by MPD Intraductal Papillary Mucinous Neoplasm (IPMN) or mix type IPMN.}
\label{tab:characteristics}
\begin{tabular}{lcc}
\hline
\rowcolor{gray!20} \textbf{Characteristics} & \textbf{Development Dataset} & \textbf{Test Dataset} \\ \hline
Number of MRCP volumes & 201 & 55 \\
\rowcolor{gray!20} Patient demographics & & \\
\hspace{1em} Number of individuals & 55 & 14 \\ 
\rowcolor{gray!20}  \hspace{1em} Age, mean (SD), years & 72.4 (9.3) & 75.6 (7.9) \\
\hspace{1em} Male : Female & 44 (80.0\%) : 11 (20.0\%) & 5 (35.7\%) : 9 (64.3\%) \\
\rowcolor{gray!20}   \hspace{1em} Normal MPD & 44 (80.0\%) & 10 (71.4\%) \\
\hspace{1em} Dilated MPD due to mixed type IPMN & 10 (18.1\%) & 3 (21.4\%) \\
\rowcolor{gray!20}  \hspace{1em} Dilated MPD due to MPD IPMN & 1 (1.8\%) & 1 (7.1\%) \\
\hline
\end{tabular}
\end{table}

Each input MRCP volume was denoted by $x \in \mathbb{R}^{1 \times D \times H \times W}$, where \(D\), \(H\), and \(W\) denoted the depth, height, and width of the 3D MRCP volume. The corresponding expert annotation is a binary label map $y \in \{0,1\}^{D \times H \times W}$, where \(y(v)=1\) indicates MPD and \(y(v)=0\) indicates background. The MPD is a tubular structure with varying width, and in terms of the imaged MRCP volume, only consists of a small fraction of the full MRCP volume ($0.019\%$). Therefore, a direct segmentation approach is affected by a severe class imbalance. Instead, we adopt a coarse-to-fine strategy: a coarse model first localises a high-recall candidate MPD region, and a fine model then performs detailed segmentation inside the cropped full resolution ROI.

\subsection{Segmentation models}

We evaluated established 3D segmentation models rather than proposing a new backbone architecture. The purpose of the model comparison was to identify a reliable and feasible model for MPD segmentation and downstream maximum-width estimation from 3D MRCP. The final selected model used a 3D U-Net as the coarse localisation model and a 3D SegResNet as the fine segmentation model.

Model selection criteria include: (1) practical training and inference GPU memory demand, (2) stable convergence on the collected dataset, (3) segmentation performance on validation set. 

\paragraph{3D U-Net}
A 3D U-Net~\cite{cciccek20163d} was used as a convolutional encoder--decoder baseline from MONAI, and selected as the coarse localisation model with full 3D MRCP image volumes as input. The model was implemented using the MONAI \texttt{UNet} module~\cite{cardoso2022monai} and the 
input and output channels were adapted to single-channel MRCP input and binary MPD segmentation.

\paragraph{Swin UNETR}
Swin UNETR~\cite{hatamizadeh2021swin} was used as a transformer-based volumetric segmentation baseline from MONAI with \texttt{SwinUNETR} backbone from MONAI~\cite{cardoso2022monai} . We used the sliding window input shape of $64 \times 64 \times 64$, embedding dimension 12, depths $(1,1,2,1)$, and attention heads $(3,6,12,24)$. The input and output channels were adapted to single-channel MRCP input and binary MPD segmentation. No task-specific architectural modifications were introduced.

\paragraph{DynUNet}
DynUNet~\cite{cardoso2022monai} is a dynamic UNet model heavily based on nnU-Net~\cite{isensee2021nnu}, which was selected as a self-configuring medical image segmentation pipeline. We used \texttt{DynUNet} from MONAI with the 3D full-resolution configuration. The framework automatically determined the patch size, batch size, network topology, and training schedule from the training data.

\paragraph{Proposed coarse-to-fine pipeline}
The proposed pipeline included two stages, where the first stage includes a coarse 3D U-Net segmentation model that receives a full MRCP volume at four-times downsampled spatial resolution with the goal of approximate localisation of the MPD duct with near-perfect recall. This additional stage is important because the MPD occupies a very small fraction of the full image volume, making direct full-resolution segmentation inefficient and highly imbalanced. The data sample is defined as: $x^{c} = \mathcal{D}(x)$, where $x^{c} \in \mathbb{R}^{1 \times D_c \times H_c \times W_c}$ and $p^{c} = f_{\theta_c}(x^{c})$, and the prediction is defined as: $p^{c} \in [0,1]^{1 \times D_c \times H_c \times W_c}$. Given a downsampled MRCP volume, the coarse 3D U-Net predicts a binary MPD segmentation map over the whole 3D image volume. The segmentation prediction on the downsampled 3D image volume is then used to crop the region of interest, defined as: $x_{\mathrm{roi}} = \mathcal{C}(x, b)$, where $x_{\mathrm{roi}} \in \mathbb{R}^{1 \times d \times h \times w}$. To avoid missing potential duct voxels during the ROI cropping stage, the ROI is spatially expanded by a fixed safety margin in all directions. In practice, we set the safety margin to 10. The resulting bounding box is used to crop a full-resolution ROI from the original MRCP volume ROI expanded with a safety margin. In this way, the fine model receives an input that is much smaller than the full scan but still contains the full MPD region with sufficient surrounding context.

In the second stage, a fine-resolution 3D SegResNet was applied to the cropped ROI. Since the search space had already been reduced by the coarse localisation stage, the fine model could focus on detailed MPD boundary delineation instead of global duct detection. This improved the effective foreground-to-background ratio and allowed the fine model to use higher spatial resolution around the duct. Final prediction is defined as: $p^{f} = f_{\theta_f}(x_{\mathrm{roi}})$, where $p^{f} \in [0,1]^{1 \times d \times h \times w}$. The fine model predicts the final binary MPD segmentation inside the cropped ROI. During inference, this ROI-level prediction is mapped back to the original image space using the crop coordinates obtained from the coarse stage, yielding the final full-volume segmentation result.

\subsection{Implementation details}

Both the coarse and fine segmentation models were trained using a combination of weighted cross-entropy and Dice loss:
\begin{align}
\mathcal{L}_{\mathrm{total}} =
\lambda_{\mathrm{CE}}\,\mathcal{L}_{\mathrm{CE}}(p,y)
+
\lambda_{\mathrm{Dice}}\,\mathcal{L}_{\mathrm{Dice}}(p,y),
\end{align}
in which the 
weighted cross-entropy loss is defined as 
\begin{equation}
\mathcal{L}_{\mathrm{CE}} =
-\sum_{v}\sum_{k \in \{0,1\}} w_k\,y_k(v)\log p_k(v),
\end{equation}
where \(w_k\) is the class weight for class \(k\). The MPD class was assigned a larger weight because MPD voxels are much rarer than background voxels.
The Dice loss for the MPD class is
\begin{equation}
\mathcal{L}_{\mathrm{Dice}} =
1-
\frac{2\sum_v p_1(v)y_1(v)+\epsilon}
{\sum_v p_1(v)+\sum_v y_1(v)+\epsilon},
\end{equation}
where \(p_1(v)\) and \(y_1(v)\) denote the predicted probability and ground-truth label for MPD at voxel \(v\), and \(\epsilon\) is a small constant for numerical stability.

During training, we applied multiple 3D data augmentations, including random flips, rotations, intensity scaling, intensity shifting, and additive Gaussian noise. All experiments were performed on NVIDIA H200 GPUs with 141\,GB memory.

\paragraph{Five-fold cross-validation ensemble}
To obtain the final segmentation prediction and enable uncertainty estimation, we trained a five-model ensemble using 
five-fold cross-validation at the patient-level on the training set. The test set was kept separate and was not used during model training.
During inference on the test set, each of the five model pairs was applied independently to every test case. The five probability maps were then averaged to form the ensemble prediction as follows:
\begin{equation}
\bar{p}(v) = \frac{1}{5}\sum_{j=1}^{5} p^{(j)}(v),
\end{equation}
where \(p^{(j)}(v)\) is the MPD probability predicted by the \(j\)-th model at voxel \(v\). The final segmentation was calculated by thresholding \(\bar{p}(v)\). In addition, the predictions of five models were further used in the uncertainty analysis described next.
The final ensemble segmentation is obtained by thresholding the ensemble mean:
\begin{equation}
\hat{y}_{\mathrm{ens}}(v)=\mathbf{1}\bigl(\bar{p}(v)\geq 0.5 \bigr).
\end{equation}

\paragraph{Uncertainty estimation}

The uncertainty of deep learning models were used to assess their utility by analysing the relationship and correlation between the uncertainty and the performance of the model. To improve model calibration, we trained an ensemble of five coarse-to-fine model pairs using five-fold cross-validation on the development set. In each fold, four folds were used for model training and one fold was used for model validation. Model performance on unseen data was evaluated using a test that was not used during any of the model development. During inference, the five trained coarse-to-fine models were applied independently to each test case producing five independent prediction masks. The prediction masks were then aggregated to produce the final segmentation and voxel-wise uncertainty as shown next.

The uncertainty of the model 
at the voxel \(v\) can be estimated using the predictive entropy of the ensemble prediction:
\begin{equation}
H(v)=
-\bar{p}(v)\log\bigl(\bar{p}(v)+\epsilon\bigr)
-
\bigl(1-\bar{p}(v)\bigr)\log\bigl(1-\bar{p}(v)+\epsilon\bigr).
\end{equation}
We evaluated the uncertainty using volume-level aggregation with three commonly applied measures derived only from model output. A straightforward way to produce a volume-level uncertainty, is to take average of all voxel-level predictive entropy values:
\begin{equation}
H(V) = \sum_{v \in V} \frac{H(v)}{|V|},
\end{equation}
where $|V|$ is the total count of voxels of volume $V$. Another aggregation measure only inspects the positively predicted voxels with so called structure-based predictive entropy:
\begin{equation}
    SH(V) = \frac{\sum_{v \in V} \hat{y}_{\mathrm{ens}}(v) \cdot H(v)}{\sum_{v \in V} \hat{y}_{\mathrm{ens}}(v)}.
\end{equation}
A third commonly used model uncertainty estimate is the coefficient of variation measure, which is defined by the dispersion and mean value of model predictions in the ensemble as follows:
\begin{equation}
CV(V) = \frac{\sigma_g}{\mu_g} = \frac{\sqrt{\frac{1}{M}\sum_{m=1}^{M}\left(|V_m| - \frac{1}{M}\sum_{m=1}^{M} |V_m| \right)^2}}{\frac{1}{M}\sum_{m=1}^{M} |V_m|},
\end{equation}
where $|V_m| = \sum_{v \in V} \mathbf{1}\bigl(p_m(v) \geq 0.5\bigr)$ for model $m \in M$.

\paragraph{Post-processing}
To improve robustness, a connected-component analysis is applied to the final prediction mask to remove small and isolated false-positive regions that are not connected to the largest component that is expected to be the MPD. This post-processing step reduces orphan positive voxels that are more likely to be false positive, thus improving the performance of downstream MPD maximum-width analysis.

\subsection{Evaluation metrics}

The performance of the model 
was evaluated in terms of the segmentation performance using volumetric similarity and surface distance measures. In addition, two abdominal radiologists performed a qualitative evaluation of 
segmentation performance as well as image quality and noted clinically relevant findings. Moreover, in this section we defined how the uncertainty of the model was quantified.

\paragraph{Segmentation evaluation}
The Dice similarity coefficient, precision and recall were defined as:
\begin{equation}
\mathrm{DSC}=\frac{2\,\text{TP}}{2\,\text{TP}+\,\text{FP}+\,\text{FN}},
\qquad
\mathrm{Precision}=\frac{\text{TP}}{\text{TP}+\,\text{FP}},
\qquad
\mathrm{Recall}=\frac{\text{TP}}{\text{TP}+\,\text{FN}},
\end{equation}
where TP, FP, and FN denote true-positive, false-positive, and false-negative MPD voxels, respectively. 
The surface distance was evaluated using average symmetric surface distance (ASSD) and 95th percentile Hausdorff distance (HD95). Let \(S_{\mathrm{pred}}\) and \(S_{\mathrm{ref}}\) denote the predicted mask and ground truth mask, and let \(d(a,S)=\min_{b\in S}\|a-b\|_2\) show the shortest distance between a point $a$ and a surface $S$. ASSD was defined as:
\begin{equation}
\mathrm{ASSD} = \frac{
\sum_{a\in S_{\mathrm{pred}}}d(a,S_{\mathrm{ref}})+
\sum_{b\in S_{\mathrm{ref}}}d(b,S_{\mathrm{pred}})}{|S_{\mathrm{pred}}|+|S_{\mathrm{ref}}|},
\end{equation}
and HD95 was defined as the 95th percentile of the pooled bidirectional surface distances:
\begin{equation}
\mathrm{HD95}=P_{95}\left(
\{d(a,S_{\mathrm{ref}}):a\in S_{\mathrm{pred}}\}
\cup
\{d(b,S_{\mathrm{pred}}):b\in S_{\mathrm{ref}}\}
\right).
\end{equation}

\paragraph{Qualitative evaluation}

The test was independently reviewed by two abdominal radiologists. For each case, the readers recorded the maximum MPD width from the reference segmentation and from the model prediction. The readers also assigned a visual quality score to the prediction of the model 
and to the original MRCP images using a five-point Likert scale and provided reasoning for the grading. The Likert for model prediction are as follows: 1: Poor (not clinically usable at all), 2: Low, some usable parts mostly not usable (large errors, fine details do not matter), 3: Medium, many usable parts but some parts problematic (fine details missing), 4: Good, almost fully usable, some minor issues, 5: Perfect, fully usable, no major or minor issues. The Likert for the MRCP image quality are defined as follows: 1. Very poor, 2. Poor, 3. Fair, 4. Good, 5. Excellent.

Model performance in terms of the max MPD width was assessed by averaging the two readers' measurements into a consensus and evaluated with mean absolute error (MAE) between the ground truth and model prediction. We further evaluated whether the model-derived maximum MPD width could support binary categorisation of duct dilatation. The cases were divided into two groups using a maximum-width threshold of \(5\,\mathrm{mm}\): non-dilated MPD (\(<5\,\mathrm{mm}\)) and dilated MPD (\(\geq5\,\mathrm{mm}\)). 
using MAE and accuracy metrics. 

\paragraph{Uncertainty evaluation}

The utility of the model uncertainty by the model was evaluated using referral-based analysis and correlation analysis.  In the referral-based analysis, each sample is ordered by an uncertainty measure and then removed from the dataset for expert referral. The mean qualitative and quantitative performances of the remainder is then calculated, producing a lineplot of expected fully-automated performance if a proportion of the data is referred. We evaluate the referral from 100\% to 30\%, as the noise in the results at the lower count of samples increases considerably. In correlation analysis, quantitative measures are directly compared to an uncertainty measure in a scatterplot, and a linear regression model is fitted to show the correlation coefficient and statistical significance. In addition, two radiologists' mean values for each of the quantitative measures are evaluated against an uncertainty measure using boxplots for each of their ordinal categories, which can be visually evaluated for correlation. Moreover, we calculate the Pearson correlation coefficient ($r$) when comparing the uncertainty measure and the segmentation metrics, and the Spearman rank correlation coefficient ($\rho$), when comparing the uncertainty measure and qualitative metrics.

\section{Results}

\subsection{Segmentation performance}

The segmentation performance was evaluated and reported for both coarse and fine models with a more detailed analysis of the final segmentation performance of 
the fine models.


\paragraph{Coarse model} 3D U-Net achieved near-perfect ROI recall as the coarse model, which means that nearly all of the reference MPD voxels were included within the cropped region of interest. The ROI cropping reduced the spatial size of the input for the fine segmentation stage by approximately four-fold compared with the full MRCP volume, while preserving basically full MPD coverage.

\paragraph{Pipeline} Within the coarse-to-fine framework, three fine-stage segmentation models were evaluated: 3D U-Net, Swin UNETR, and 3D SegResNet, in addition to which
the DynUNet was included as a one-stage full-volume reference pipeline. Table~\ref{tab:segmentation_results_fine} summarised the MPD segmentation performance on the test set with all four evaluated pipelines. The 3D U-Net + 3D SegResNet approach turned out to have the highest Dice similarity coefficient, precision, recall among the compared methods. These results suggest that ROI-based coarse-to-fine segmentation pipeline is well suited for MPD segmentation due to the fact the duct is a small tubular structure that occupies only a small fraction of the full MRCP volume.

\begin{table}[ht]
\centering
\caption{Main pancreatic duct segmentation performance on the test set volumes (N=55). Best results are shown in bold.}
\label{tab:segmentation_results_fine}
\begin{tabular}{lrrrrrrr}
\toprule
Method & DSC $\uparrow$ & Precision $\uparrow$ & Recall $\uparrow$ & ASSD (mm) $\downarrow$ & HD95 (mm) $\downarrow$ & Zero prediction mask $\downarrow$ \\
\midrule
DynUNet & 0.579 & 0.613 & 0.583 & \textbf{3.50} & 23.7  & \textbf{0} (\textbf{0.00\%})\\
3D U-Net + 3D U-Net &  0.629 & 0.717 & 0.585 & 4.65 & 32.0  & \textbf{0} (\textbf{0.00\%)} \\
3D U-Net + Swin UNETR &  0.516 & 0.634 & 0.479 & 11.7 & 45.5 & 1 (1.82\%) \\
3D U-Net + 3D SegResNet &  \textbf{0.634} & \textbf{0.744} & \textbf{0.593} & 4.18 & \textbf{22.9} & 1 (1.82\%)\\
\bottomrule
\end{tabular}
\end{table}

\subsection{Qualitative evaluation}

In this subsection, 
we report all the qualitative evaluation measures.

\paragraph{Max MPD Width Measurement Performance}  The MPD Max width was qualitatively measured from the segmentation model prediction and ground truth by the two radiologists using 3D Slicer. The MPD was considered dilated if the maximum width is $\geq 5$mm. The consensus of the two radiologist was defined by taking the mean of their evaluated width values. In terms of model performance compared
to ground truth, the mean absolute error was 0.355~mm. This resulted in a dilated MPD accuracy of 74.5\%, with a sensitivity of 75.0\%, a specificity of 74.4\%, a precision of 45.0\% and an F1-score of 56.3\%. For the ground truth 
annotations, the inter-rater agreement between the two radiologists was 0.618~mm MAE and the accuracy of the classification of 
DMPD was 83.6\%.

\paragraph*{Qualitative vs. Quantitative metrics}

In the Supplementary, 
we present a comparison of qualitative Likert measures for model segmentation and MRCP image quality against quantitative metrics (DSC, ASSD, HD95). In addition, the reasons for Likert grading based on radiologists' assessment are summarised in the Supplementary.

\subsection{Model Uncertainty Analysis}

\begin{figure}[!h]
    \centering
    \includegraphics[width=0.95\textwidth]{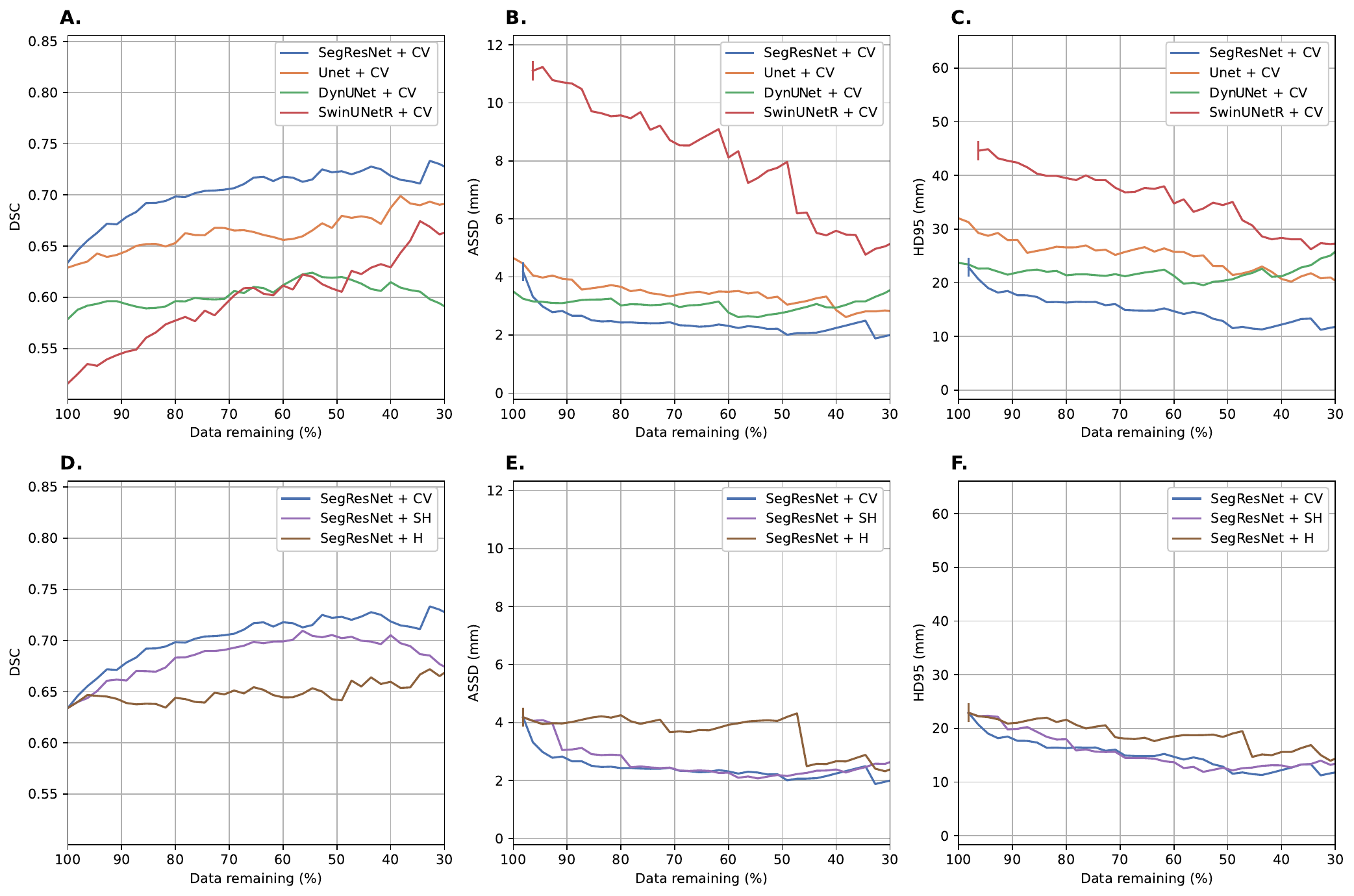}
    \caption{Model performance across uncertainty threshold on referral based approach. The most uncertain cases are referred and average performance on remaining cases is plotted. In subfigures \textbf{A.}~to~\textbf{C.} the pipelines are compared using the best-performing uncertainty measure coefficient of variation (CV), and in subfigures \textbf{D.}~to~\textbf{F.} the three uncertainty measures predictive entropy (H), structure-based predictive entropy (SH), and coefficient of variation (CV), are compared using the best performing model i.e., 3D U-Net + 3D SegResNet. Note that evaluation for two of the pipelines start at 98\% data remaining on the distance based measures due to zero prediction mask.}
    \label{fig:unc_referral}
\end{figure}

In the referral-based evaluation of the model uncertainty, we measured the performance using quantitative metrics of DSC, ASSD, and HD95. When examining the change in the DSC values at 100\% and 30\% data remaining using the coefficient of variation (CV) uncertainty measure to refer to the most uncertain cases, the results turn out to be 
as follows: DynUNet achieved 0.579 and 0.594; 3D U-Net + 3D U-Net achieved 0.629 and 0.690; 3D U-Net + Swin UNETR achieved 0.516 and 0.661; and 3D U-Net + 3D SegResNet achieved 0.634 and 0.730.
Focusing only on the best-performing model (3D U-Net + 3D SegResNet) and evaluating referral performance with the three uncertainty measures (CV, SH, and H), the initial performance at 100\% data was 0.634 DSC. The mean DSC at 60\% data remaining was 0.718 (CV), 0.699 (SH), and 0.644 (H). At 30\% data remaining, the DSC values were 0.730 (CV), 0.677 (SH), and 0.665 (H). The complete evaluation is shown in Figure~\ref{fig:unc_referral}.

\begin{figure}[!h]
    \centering
    \includegraphics[width=0.95\textwidth]{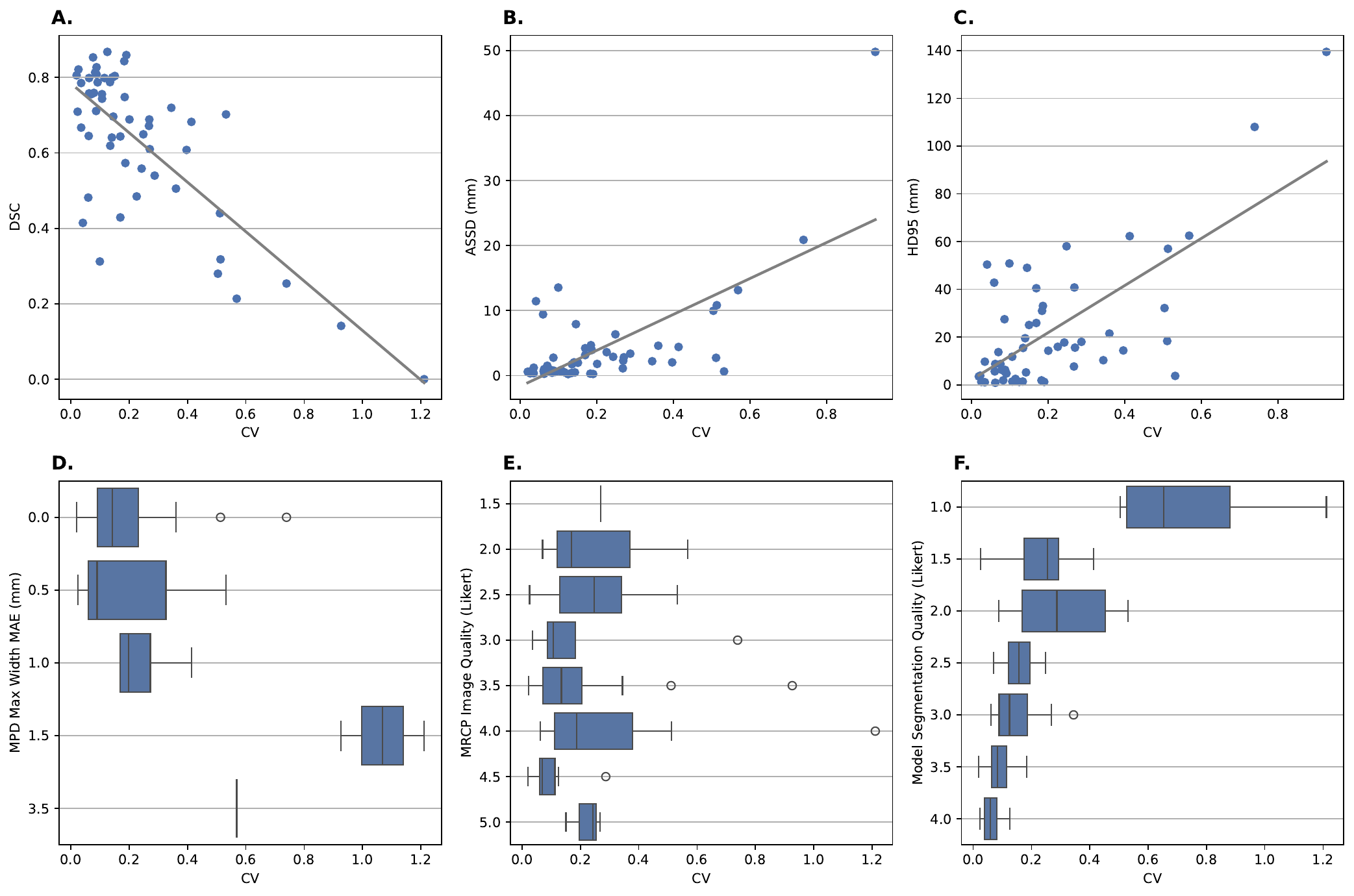}
    \caption{Correlation analysis between uncertainty measure and model performance. Subfigures \textbf{A.}~to~\textbf{C.} are scatterplots with fitted linear line between quantitative measures and coefficient of variation (CV) and subfigures \textbf{D.}~to~\textbf{F.} are boxplots with ordinal qualitative measures and CV.}
    \label{fig:unc_scatter_boxplot}
\end{figure}

In terms of correlation analysis, fitting a linear model to the CV uncertainty measure against quantitative metrics of DSC, ASSD, and HD95 showed statistically significant correlation ($p<0.001$) with strong coefficient values of $r=-0.759$, $r=0.693$ and $r=0.695$, respectively. When the CV was compared with 
qualitative measures of model performance (MPD Max Width MAE~[mm] and 
Likert scales), the correlation coefficient $\rho=0.187$ ($p=0.173$), $\rho=-0.083$ ($p=0.547$), $\rho=-0.728$ ($p<0.001$) for MPD Max Width MAE, MRCP image quality Likert, and model segmentation quality Likert, respectively. Specifically, in MPD max width MAE, the mean CV values 
were 0.18, 0.19, 0.24, 1.07, and 0.57 for the MAE values of 0, 0.5, 1.0, 1.5, and 3.5~mm, respectively. The MRCP image quality Likert did not strongly correlate with CV, having mean values of  0.27, 0.27, 0.25, 0.23, 0.20, 0.29, 0.11, and 0.22 for Likert scores of 1.5, 2.0, 2.5, 3.0, 3.5, 4.0, 4.5, and 5.0 respectively. The model segmentation quality Likert scale showed a stronger relationship with CV values of 0.74, 0.24, 0.31, 0.16, 0.16, 0.09, and 0.06 for Likert scores of 1.0, 1.5, 2.0, 2.5, 3.0, 3.5, and 4.0, respectively. Complete 
analysis is shown in Figure~\ref{fig:unc_scatter_boxplot}.

\subsection{Within-patient analysis}

\begin{figure}[!h]
    \centering
    \includegraphics[width=0.95\textwidth]{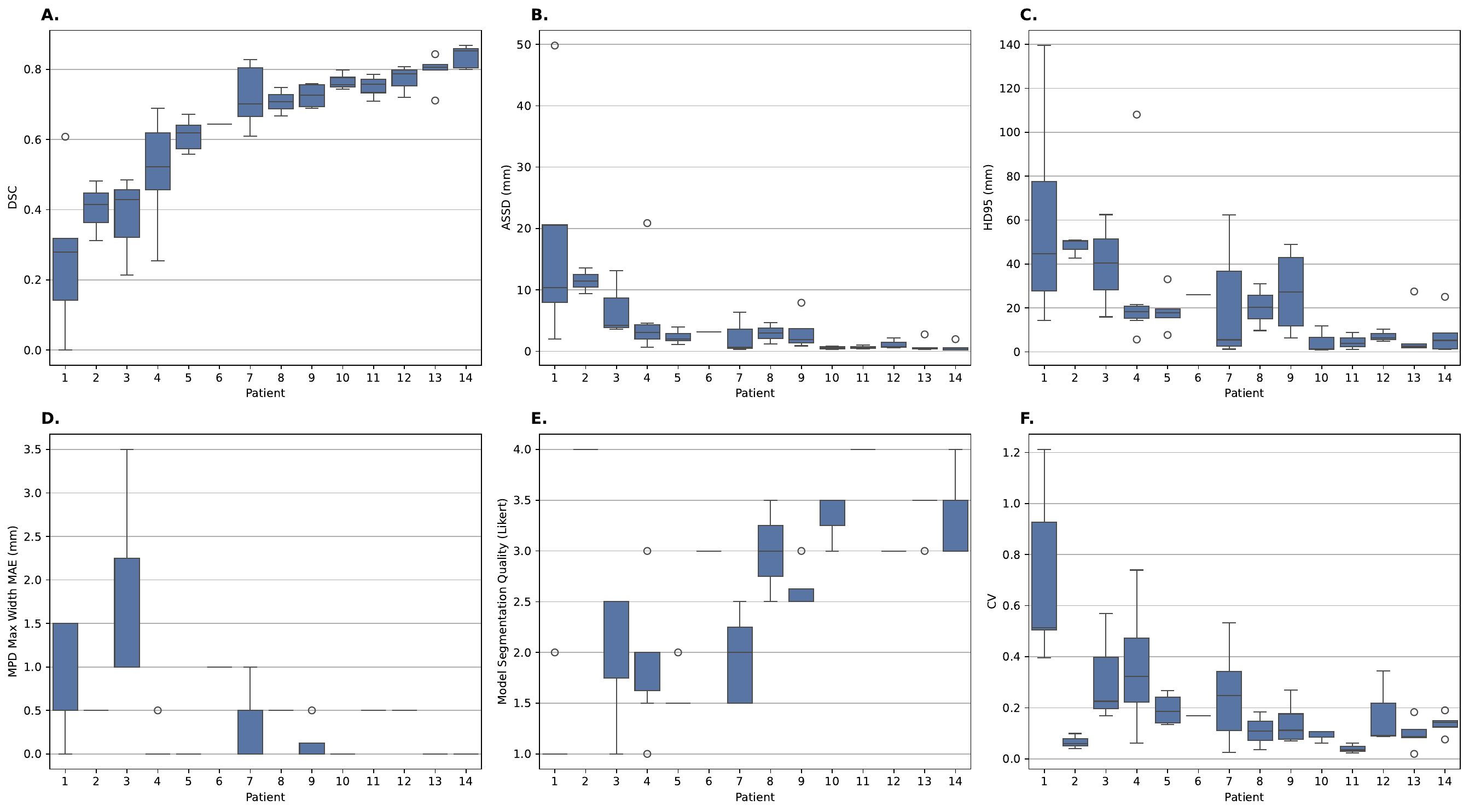}
    \caption{Detailed look at 3D U-Net + 3D SegResNet performance within each test-set patient ordered ascending by median DSC performance. Segmentation performance is shown 
    on subfigures \textbf{A.}~to~\textbf{C.} while subfigures \textbf{D.}~and~\textbf{E.} show qualitative measures by radiologists. Subfigure \textbf{F.} shows the coefficient of variation uncertainty measure.}
    \label{fig:within_patient}
\end{figure}

The best performing 3D U-Net + 3D SegResNet model and the best performing CV uncertainty measure based on scan-level analysis were inspected in more detail in terms of within-patient analysis. In this analysis, the patient-level variability is analysed for segmentation, qualitative and uncertainty estimation. In terms of variability, the mean and standard deviation distance from the patient mean was 0.05 (0.04) with DSC, 1.53 (1.55) with ASSD, 9.69 (8.16) with HD95, and 0.16 (0.32) with MPD Max Width MAE, 0.24 (0.21) with model prediction Likert quality, and 0.09 (0.08) with CV. The results are visualised in boxplots for each patient, and patients are ordered
by median DSC performance. The complete results are shown in Figure~\ref{fig:within_patient}.

\clearpage
\section{Discussion}


In terms of segmentation performance, the 3D U-Net + 3D SegResNet approach outperformed all other methods 
due to its residual convolutional feature extraction mechanism, which can 
provide stable optimization and better preservation of local ductal boundary information on a limited dataset. However, our approach 
did produce a single empty prediction mask and had the second lowest ASSD value of the compared methods. The only transformer 
based approach, i.e., Swin UNETR, had the lowest performance out of all compared methods. This can likely be attributed to the relatively small
amount of data for the transformer-based approach.

In the downstream task of MPD max width estimation and dilated MPD classification from deep learning-based segmentation masks, the automatic approach achieved good performance, though still with lower value 
than the inter-rater agreement, which is consistent with the segmentation metrics thus having room for improvement. The sensitivity indicates that most of the dilated cases were detected, while the relatively low precision reflects false-positive dilatation predictions. These results suggest that the model-derived maximum width may be useful for screening or prioritisation, but false-positive cases would still require review by a radiologist.

In qualitative analysis, the mean visual evaluation of the clinical usability of the prediction of the model using the Likert grade was 2.67 by the first radiologist and 2.50 by the second radiologists. 
This indicates that the predicted segmentations were often imperfect as complete anatomical masks. However, the sub-millimeter maximum-width error suggests that the method may still be useful as a measurement-support tool, especially when the intended endpoint is maximum MPD width rather than complete duct delineation. An important finding was that the prediction errors did not exhibit a readily identifiable pattern. In a single case, the model failed to detect MPD completely, producing a clinically misleading measurement, and it occasionally produced subtle and visually conspicuous ductal abnormalities, and some predictions extended outside the ductal lumen. This suggests that the performance of the model was not solely determined by the apparent complexity of the finding. One possible explanation is A shift in data distribution between training and validation datasets, including variations in anatomy, imaging protocols, or image quality, which may have affected model generalisability. This heterogeneity of failure modes complicates the identification of a single technical explanation and highlights the need for further investigation into the factors that could affect model robustness. Moreover, when comparing both segmentation quality and MRCP image quality Likert values to model performance, a visual relationship is evident for segmentation measures, but less clear for MPD max width estimation.

In the uncertainty analysis, UNet 3D + SegResNet3D had the highest performance in terms of referral-based analysis. In terms of uncertainty measures, the coefficient of variation measure outperformed the two entropy-based uncertainty measures, which is consistent with previous 
work in medical image segmentation~\cite{sahlsten2024application}. The correlation analysis demonstrated that uncertainty can estimate segmentation quality. The boxplots show visual ordinal relationships between CV and qualitative measures. In terms of DMPD classification, the model had 74.5\% accuracy, which is lower but relatively close to the inter-observer variability, where the raters had 83.6\% agreement, setting a realistic upper bound for the model performance.

There are limitations in our study related to the number of cases studied and the tools for qualitative measurement. The number of individual patients and the number of image volumes were relatively low; there are 14 patients with 55 scans included in the test set. However, the average duration between scans was nearly three years, providing more heterogeneous cases even from the same patients with generalizability.
In addition, the data were from a single hospital, which may reduce the generalizability of the results.
The reference measurements were obtained using the native measurement tool provided by 3D Slicer. As 
measurement markers partially obscure the duct margins during placement, exact identification of the measurement endpoints is inherently limited by image resolution and observer judgment. Therefore, a small degree of measurement uncertainty should be assumed when interpreting the reported distances. This partially explains the disagreement between the two raters.

Future work should focus on external validation across institutions and hospitals, as the current study is based on a single hospital. Furthermore,  
the collected dataset should scale up to improve generalizability. Methodologically, future models should improve MPD continuity, reduce zero-prediction cases, and better preserve the MPD morphology. In addition, more robust automatic width-estimation methods, such as orthogonal cross-sectional analysis, is expected to reduce dependence on manual expert measurement.

This study demonstrates the feasibility of automated segmentation of the main pancreatic duct from 3D MRCP using convolutional neural networks. Although 
segmentation of the duct remains challenging due to the MPD occupying only a small fraction of the image volume and its 
irregular morphology, the proposed pipeline supports a quantitative assessment of its maximum-width. 
Ensemble-based uncertainty analysis showed a clear correlation between higher model uncertainty and lower segmentation performance, indicating that uncertainty estimates may help identify cases that require radiologist review.
Expert validation further suggested that MRCP image quality is related to maximum-width measurement reliability, with lower image-quality scores associated with higher MPD max width measurement errors. Overall, the method shows potential as a support tool for MPD width assessment, but further external validation and robustness improvements are needed before clinical deployment.

\section*{Acknowledgments}

This research was funded by Business Finland under project Medical AI and Immersion (decision number 10912/31/2022). The funders had no role in study design, data collection and analysis, decision to publish, or preparation of the manuscript.

The authors thank Mervi Asumaniemi, ICT manager at Tampere University Hospital for her assistance in data anonymisation and data transfer. 

\section*{Author contributions}

Z.Z. contributed to conceptualisation, study design, methodology development, implementation, formal analysis, and writing of the main manuscript. J.S. contributed to study design, data collection, results analysis, figure plotting, and writing of the manuscript. M.S. contributed to results analysis, and writing of the manuscript. J.R. contributed to clinical supervision, supervised data annotation, conducted the clinical validation of the model outputs, and contributed to manuscript editing. M.L. and A.A. contributed to the data annotation. I.R. contributed to clinical supervision, supervised data annotation, conducted the clinical validation of the model outputs, and contributed to manuscript editing. K.K. contributed to conceptualisation and manuscript editing. All authors reviewed the manuscript.

\section*{Competing interest}
The authors have declared that there are no competing interests.

\section*{Data availability}
The patient data cannot be shared publicly because of the data protection law of Finland, the General Data Protection Regulation (GDPR) of European Union, and our research permission granted by Tampere University Hospital that do not allow sharing of individual patients' data. The patient data is available from Tampere University Hospital for researchers who meet the criteria for access to confidential data. More specifically the datasets are not publicly available and restrictions apply to their use. The dataset restrictions are imposed by the Wellbeing Services County of Pirkanmaa, Tampere University Hospital, Research services, P.O. Box 2000, 33521 Tampere, Finland (requests for the data access can be made to clinicaltrials@pirha.fi).

\section*{Code Availability}
The code used for data processing, deep learning algorithms, and model evaluation is not publicly available. The Methods section provides sufficient detail for replication using open-source libraries. Code is available from the corresponding author upon reasonable request.


\bibliography{refs}



\end{document}